\documentclass[11pt]{article}

\usepackage[table]{xcolor}

\usepackage[preprint]{acl}

\usepackage{times}
\usepackage{latexsym}
\usepackage{amsmath}
\usepackage{amssymb}
\usepackage{booktabs}
\usepackage{float}
\usepackage{hyperref}
\usepackage{algorithm}
\usepackage{multirow}
\usepackage[noend]{algpseudocode}
\usepackage{subcaption}
\usepackage{enumitem}
\definecolor{mylightblue}{rgb}{0.8, 0.9, 1.0}
\usepackage{siunitx}

\usepackage[T1]{fontenc}
\usepackage[utf8]{inputenc}

\usepackage{microtype}

\usepackage{inconsolata}

\usepackage{graphicx}

\title{ThinkSwitcher: When to Think Hard, When to Think Fast}

\author{
Guosheng Liang, Longguang Zhong, Ziyi Yang, Xiaojun Quan\thanks{Corresponding author}\\
School of Computer Science and Engineering, Sun Yat-sen University \\
\texttt{\{lianggsh3,zhonglg5,yangzy39\}@mail2.sysu.edu.cn}\\
\texttt{quanxj3@mail.sysu.edu.cn}\\ 
}

\begin{document}
\maketitle
\begin{abstract}
Large reasoning models (LRMs) excel at solving complex tasks by leveraging long chain-of-thought (CoT) reasoning. However, this often leads to overthinking on simple tasks, resulting in unnecessary computational overhead. We observe that LRMs inherently possess the capability for efficient short CoT reasoning, which can be reliably elicited through prompt design. To leverage this capability, we propose \textbf{ThinkSwitcher}, a framework that enables a single LRM to dynamically switch between short and long CoT modes based on task complexity. ThinkSwitcher introduces a lightweight switching module trained with supervision signals derived from the relative performance of each reasoning mode across tasks. Experiments on multiple reasoning benchmarks show that ThinkSwitcher reduces computational cost by 20–30\% while maintaining high accuracy on complex tasks. This demonstrates the effectiveness of ThinkSwitcher as a scalable and efficient solution for unified LRM deployment.
\end{abstract}

\vspace{0.1cm}
\section{Introduction}
Large reasoning models (LRMs) \cite{openai2024openaio1card, guo2025deepseek, google_gemini_25_pro, Claude3S} have demonstrated impressive capabilities in solving complex tasks. They achieve this through long chain-of-thought (CoT) processes, which involve behaviors such as exploration, self-reflection, and verification \cite{li2025system,gandhi2025cognitive,zeng2025simplerl}.
However, this strength can become a drawback: LRMs tend to \textit{overthink} \cite{chen2024not, sui2025stop, cuadron2025danger} even the simplest problems (e.g., \textit{2+3=?}), which unnecessarily invokes elaborate reasoning for tasks that require minimal effort. This inefficiency becomes especially problematic in high-throughput applications \cite{kumar2025overthink, qu2025survey}. To mitigate this problem, many systems deploy two separate models: one tailored for complex reasoning and another for simpler tasks. While effective, this dual-model setup incurs additional computational and memory costs. 
This raises a fundamental question: \textit{Can a single model achieve both robust reasoning capabilities and high efficiency?}

\begin{figure}[!t]
\centering
\vspace{-0.2cm}
\includegraphics[width=0.48\textwidth]{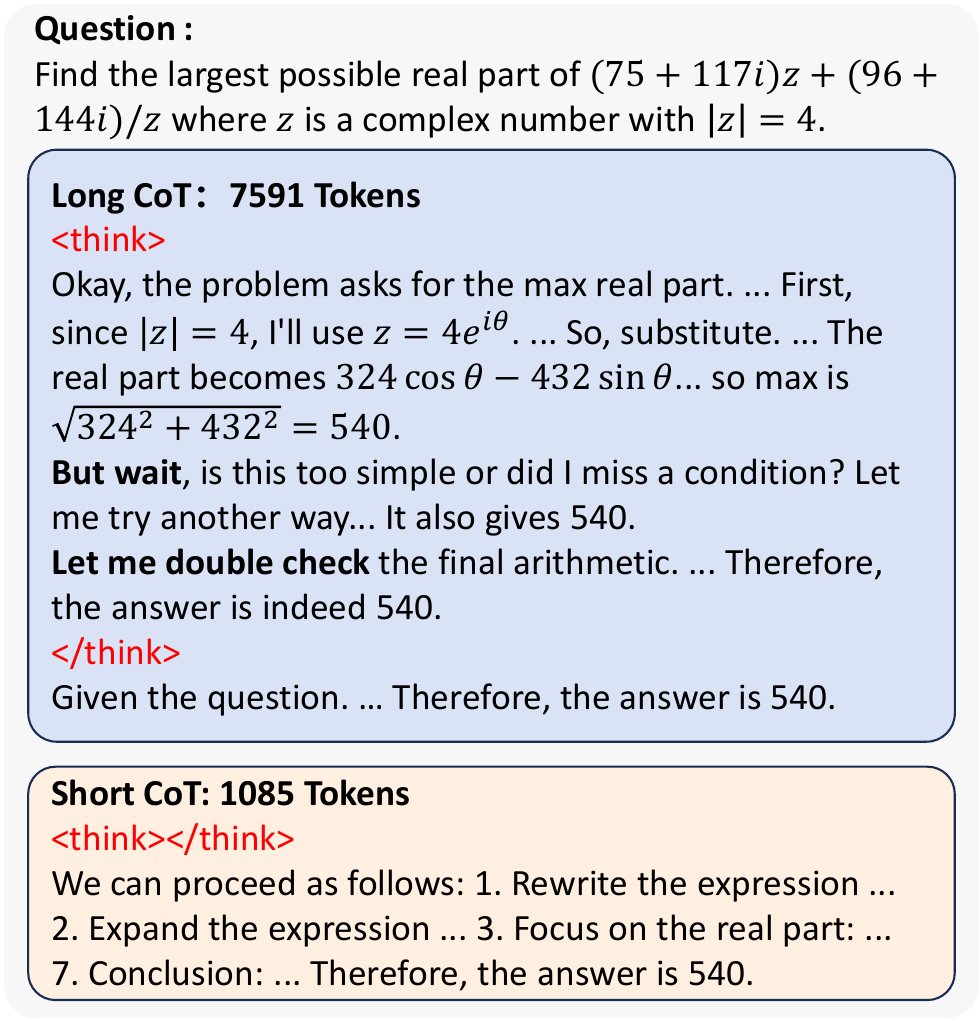}
\caption{Comparison of long and short CoTs generated using different prompting strategies with Deepseek-R1-Distill-Qwen-7B. While long CoT reasoning often leads to \emph{overthinking} and excessive token consumption due to elaborate reasoning steps, the short CoT can deliver comparable accuracy with substantially fewer tokens.
}
\label{fig:Comparison_Long_Short_CoT}
\vspace{-0.4cm}
\end{figure}

Inspired by the adaptive nature of human cognition—such as the \textit{System 1} and \textit{System 2} framework \cite{kahneman2003maps,kahneman2011thinking,hua2022system}—we investigate enabling a single powerful reasoning model to operate in two distinct modes: its native long CoT mode for complex problems and an efficient short CoT mode for simpler tasks. Recent works such as Gemini-2.5-Pro \cite{google_gemini_25_pro}, Qwen3 \cite{qwen3}, and Llama-Nemotron \cite{bercovich2025llama} have also explored dual-mode systems. However, these approaches often lack public implementation details \cite{google_gemini_25_pro} or depend on post-training with curated long and short CoT data \cite{qwen3,bercovich2025llama}, and typically require manual mode selection based on user inputs. In contrast, we propose a lightweight and adaptive alternative: a switcher module that automatically selects the appropriate reasoning mode based on task complexity, without backbone changes or large-scale training.

Our investigation reveals a key insight, echoed by \citet{ma2025reasoning}: advanced LRMs already possess a latent capability for concise and effective short CoT reasoning. 
Figure~\ref{fig:Comparison_Long_Short_CoT} illustrates a case where long CoT reasoning results in excessive token consumption, while short CoT achieves an accurate answer with substantially fewer tokens. 
Notably, we find that the short CoT capability can be reliably activated by appending an empty thinking block (e.g., \textit{<think></think>}) after the user instruction—a phenomenon also observed in Qwen3~\cite{qwen3}. This simple prompt-based intervention requires no changes to the model itself. Beyond this empirical finding, we offer a theoretical explanation of this latent behavior and its supporting mechanisms, detailed in Appendix~\ref{appendix:mechanism_behind_short_cot}.

Building on these observations, we propose the \textbf{ThinkSwitcher} framework, which enables a single LRM to adaptively switch between long and short CoT modes. To support this capability, a lightweight switcher module is employed to predict the reasoning mode likely to yield optimal performance for a given query. The switcher is trained using self-supervised signals derived from the backbone model’s own performance when executing both reasoning modes. This eliminates the need for external annotation or extensive post-training.

Our experiments show that ThinkSwitcher noticeably reduces average token usage across various benchmarks while maintaining high accuracy on complex reasoning tasks by retaining long CoT where necessary. For example, on simpler datasets such as GSM8K~\cite{cobbe2021training}, it reduces inference tokens by around 30\% with a performance loss of less than 1\%. On more challenging datasets like AIME \cite{AIME}, ThinkSwitcher achieves token reductions of 38\%, with only approximately a 2\% decline in performance. Overall, it consistently lowers computational costs by 20–30\% across benchmarks while retaining highly competitive accuracy. These results validate our approach to unifying strong reasoning capabilities with efficient resource usage in a single model deployment.

\section{Related work}
\paragraph{Large Reasoning Models}  
Large reasoning models (LRMs), such as OpenAI-o1~\cite{openai2024openaio1card}, DeepSeek-R1~\cite{guo2025deepseek}, and QwQ~\cite{qwen_qwq_32b}, are designed to emulate System-2 reasoning~\cite{li2025system}. These LRMs have demonstrated state-of-the-art performance on challenging tasks in mathematics~\cite{cobbe2021training, hendrycks2measuring} and coding~\cite{chen2021evaluating, codeforces}. These models are typically trained via reinforcement learning (RL) algorithms~\cite{PPO,GRPO} to elicit long chain-of-thought (CoT) reasoning. However, their tendency to \textit{overthink}~\cite{chen2024not} even on simple questions results in substantial computational inefficiencies and restricts their practicality in high-throughput scenarios.
Several dual-system models have been introduced to mitigate this, including Claude-3.7-Sonnet~\cite{Claude3S}, Gemini-2.5-Pro~\cite{google_gemini_25_pro}, Qwen3~\cite{qwen3}, and Llama-Nemetron~\cite{bercovich2025llama}. These models offer both long and short CoT modes, allowing users to choose between deep reasoning and quick answers depending on themselves.  Among these, approaches from ~\citet{qwen3} and ~\citet{bercovich2025llama} implement this dual-mode capability through post-training on carefully curated mixtures of long and short CoT datasets.

Nonetheless, most of these models rely on manual mode selection, and lack the ability to automatically adapt the reasoning depth based on the input query.
In contrast, we propose a lightweight mechanism that allows a single model to dynamically switch between short and long CoT modes. This enables both efficiency and strong reasoning without modifying weights or requiring post-training.

\begin{figure*}[!ht]
\centering
% \vspace{-0.3cm}
\includegraphics[width=0.95\textwidth]{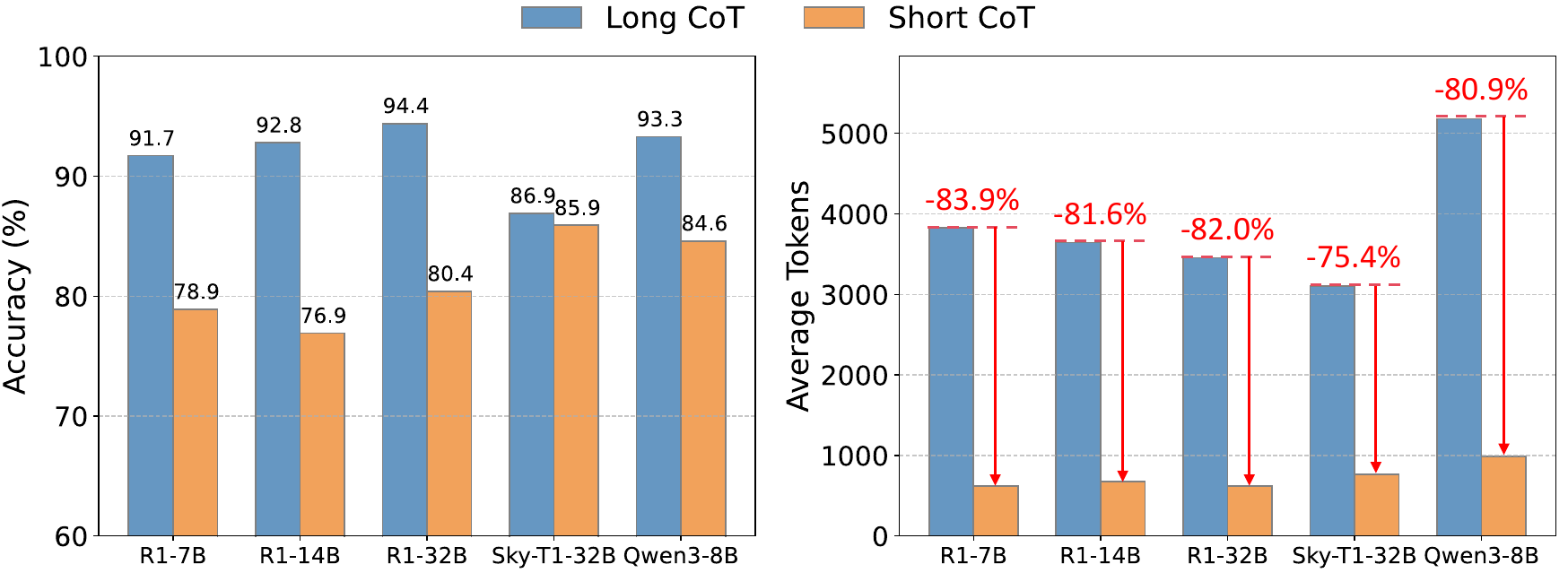}
\vspace{-0.1cm}
\caption{Comparison of long CoT and induced short CoT on the MATH500. ``R1'' denotes DeepSeek-R1-Distill series. \textbf{Left}: Accuracy comparison between long CoT and short CoT. \textbf{Right}: Average token usage for each reasoning mode, which demonstrates substantial token reductions with short CoT. Our approach of inducing short CoT consistently achieves substantial token savings while maintaining competitive accuracy across diverse LRMs.}
\label{fig:short_cot_performance}
\vspace{-0.2cm}
\end{figure*}

\paragraph{Efficient Reasoning}
Extensive research has focused on reducing inference overhead and improving reasoning efficiency \cite{sui2025stop}. Among these methods, Kimi-k1.5 \cite{team2025kimi} and O1-pruner \cite{luo2025o1} introduce length-controlled reward functions in reinforcement learning to reduce CoT reasoning length. Alternatively, methods such as DAST \cite{shen2025dastdifficultyadaptiveslowthinkinglarge}, C3oT \cite{kang2024c3ot}, and TokenSkip \cite{xia2025tokenskipcontrollablechainofthoughtcompression} train LRMs to generate compact CoTs by constructing datasets with varying reasoning lengths and applying post-training techniques like SFT, DPO \cite{rafailov2023direct}, or SimPO \cite{meng2024simpo}. Nevertheless, both strategies necessitate additional LRM training, which introduces significant computational costs.
A distinct direction involves prompt-guided strategies, such as CoD \cite{xu2025chaindraftthinkingfaster} and CCoT \cite{Renze_2024}, which guide models to directly produce concise reasoning without fine-tuning. However, these methods lack the ability to dynamically adapt reasoning depth to question complexity, leading to performance degradation on challenging tasks \cite{xu2025chaindraftthinkingfaster}. 
Another line of work on LLM-routing, such as RouteLLM \cite{ong2025routellm}, employs a router to distribute user questions across specialized LLMs. This approach assigns questions to the most suitable model to reduce average inference costs. While effective, it requires the simultaneous deployment of multiple LLMs.
In contrast, our method achieves efficient reasoning with a single LRM, avoiding both significant fine-tuning costs and multi-model deployment complexities. This orthogonal approach dynamically adjusts reasoning depth and strikes an effective balance between deployment overhead and system performance.

\vspace{-0.1cm}
\section{Observations}
\label{sec:observations}
\vspace{-0.1cm}
Large reasoning models (LRMs) such as DeepSeek-R1~\cite{guo2025deepseek}, Sky-T1-32B~\cite{sky_t1_2025}, and Qwen3~\cite{qwen3} commonly employ structured generation formats to tackle complex reasoning tasks.
A notable characteristic of these models is the use of special tokens, typically \texttt{<think>} and \texttt{</think>}, which explicitly separate the model's intermediate reasoning process from the final answer.
The content after the \texttt{</think>} token provides a concise summary of the preceding reasoning and presents the final response.

\subsection{Inducing Short CoT in LRMs}
\label{subsec:core_observation}
\vspace{-0.07cm}
Our central observation is that the reasoning behavior of these models can be substantially influenced by manipulating the content placed within the \texttt{<think>} and \texttt{</think>} delimiters. In particular, we find that minimal or suggestive prompts within the \texttt{<think>} block can effectively steer the model toward generating much shorter chains of thought—termed \textit{short CoT}—thereby countering its default tendency toward verbose reasoning.

For instance, prompts such as:
\vspace{-0.1cm}
\begin{itemize}
\item \texttt{<think>}This problem appears straightforward.\texttt{</think>} (hinting at low complexity)
\vspace{-0.5cm}
\item \texttt{<think></think>} (no explicit reasoning)
\end{itemize}
\vspace{-0.1cm}
consistently lead to more concise reasoning outputs compared to the model’s standard behavior. 
This finding suggests that the reasoning trajectories of LRMs are highly prompt-sensitive and can be guided in depth and length through carefully designed prompts, even without explicitly instructing the model to generate shorter outputs.

\vspace{-0.1cm}
\subsection{Performance of Induced Short CoT}
\label{subsec:generality_efficiency_performance}
\vspace{-0.07cm}
To evaluate the utility of prompt-induced short CoT as an efficient reasoning strategy, we conduct a series of experiments across multiple reasoning models.
We focus on two key questions:
\vspace{-0.1cm}
\begin{enumerate}[label=(\arabic*)]
    \item \textit{To what extent does short CoT reduce computational cost?}
    \vspace{-0.25cm}
    \item \textit{How much reasoning performance is preserved relative to long CoT?}
\end{enumerate}
\vspace{-0.1cm}

As shown in Figure~\ref{fig:short_cot_performance}, our method consistently yields significantly shorter outputs across all models, including DeepSeek-R1-Distill (7B, 14B, and 32B), Sky-T1-32B, and Qwen3-8B. Additional figures illustrating these trends on other benchmarks are provided in Appendix~\ref{appendix:additional_shortcot_results}. This reduction in output length implies improved computational efficiency, potentially lowering FLOPs, reducing memory usage, and accelerating inference.

While short CoT shows a moderate drop in accuracy on particularly complex problems, it remains competitive across most cases and retains strong problem-solving capability. These results indicate that short CoT can serve as an efficient default reasoning mode, with long CoT invoked only when necessary. Furthermore, theoretical analysis of the mechanisms underlying short CoT induction is provided in Appendix~\ref{appendix:mechanism_behind_short_cot}, offering a deeper explanation that extends the observations made in this section.

\vspace{-0.1cm}
\section{Methodology}
\label{sec:method}
\begin{figure}[!t]
    \centering
    \includegraphics[width=0.48\textwidth]{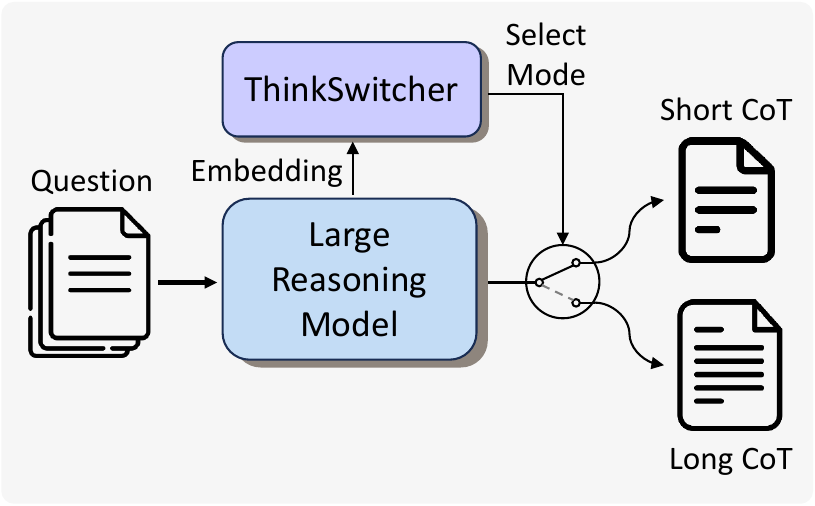}
    \caption{Dynamic mode selection during inference. Given a question embedding from the LRM, ThinkSwitcher dynamically chooses between short and long CoT reasoning based on estimated task difficulty.}
    \label{fig:thinkswitcher_framework}
\vspace{-0.3cm}
\end{figure}

\vspace{-0.1cm}
To strike an effective balance between reasoning performance and computational efficiency, we propose \textbf{ThinkSwitcher} which dynamically switches between short and long chain-of-thought reasoning modes based on the input. Figure~\ref{fig:thinkswitcher_framework} illustrates the overall workflow of this framework.

\vspace{-0.1cm}
\subsection{Data Construction}
\label{subsec:data_construction}
Training ThinkSwitcher requires reliable, fine-grained supervision signals that reflect the relative effectiveness of short and long CoT reasoning for each input question. To this end, we adopt a multi-sample evaluation strategy and construct continuous regression targets based on empirical pass rates, rather than relying on unstable single-response outcomes or coarse binary labels. The data construction process is as follows:

\textbf{Step 1: Prompting.} 
For each query $q \in \mathcal{D}_\text{train}$, we construct two prompts corresponding to short CoT (SC) and long CoT (LC) reasoning modes, denoted by $m \in \{\text{SC}, \text{LC}\}$. The specific prompt templates are provided in Appendix~\ref{appendix:prompt_templates}.

\textbf{Step 2: Sampling.}  For each query, we generate $k$ responses under both reasoning modes:
\begin{equation}
\mathcal{R}_m(q) = \left\{ r^{(i)}_m(q) \sim \pi(\cdot \mid q) \right\}_{i=1}^{k}.
\end{equation}
\textbf{Step 3: Evaluation.}  
Each response is checked for correctness. The empirical pass rate is:
\begin{equation}
\mathcal{P}_m(q) = \frac{1}{k} \sum_{i=1}^{k} \mathbb{I} \left[ r^{(i)}_m(q) \text{ is correct} \right].
\end{equation}

\textbf{Step 4: Labeling.}  The pass rate serves as the regression target for supervising the mode switcher:
\begin{equation}
y_m(q) = \mathcal{P}_m(q) .
\end{equation}
The resulting training data takes the form $(x_q, y_{\text{SC}}(q), y_{\text{LC}}(q))$, where $x_q$ is the query embedding and $y_{\text{SC}}(q), y_{\text{LC}}(q) \in [0, 1]$ denote the empirical pass rates under short and long CoT modes.

\subsection{Decision Rule for Switching}
\label{subsec:switcher_architecture}
ThinkSwitcher is implemented as a lightweight regressor that predicts the expected pass rates for short and long CoTs given a query $q$. It is trained using the supervision targets $y_{\text{SC}}(q)$ and $y_{\text{LC}}(q)$ defined in Section~\ref{subsec:data_construction}. The input is the query embedding $x_q$ extracted from the reasoning model.

At inference time, the switcher takes $x_q$ as input and produces two scalar predictions:
\begin{equation}
\left[ \hat{y}_{\text{SC}}(q),\ \hat{y}_{\text{LC}}(q) \right] = f_\phi(x_q), 
\end{equation}
which correspond to the estimated pass rates under short and long CoT prompting, respectively. The final decision is made by comparing the difference between them against a tunable threshold $\tau$:
\begin{equation}
\label{eq:decision_rule}
m(q) =
\begin{cases}
    \text{LC}, & \text{if } \hat{y}_{\text{LC}}(q) - \hat{y}_{\text{SC}}(q) \ge \tau , \\
    \text{SC}, & \text{otherwise} .
\end{cases}
\end{equation}
This means the long CoT pathway is selected when its predicted advantage over short CoT exceeds the threshold $\tau$; otherwise, the short CoT pathway is used, as formalized in Equation~(\ref{eq:decision_rule}).

\subsection{Margin-Aware Training Objective}
\label{subsec:training_loss}

We design a margin-aware objective to enhance the switcher's decision quality. A core component of this objective is the mean squared error (MSE), where $\text{MSE}(\hat{y}, y)$ denotes the squared difference $(\hat{y}-y)^2$ between a prediction $\hat{y}$ and its target $y$. The initial loss term, $\mathcal{L}_{\text{MSE}}$, applies this to the predicted and target pass rates for both reasoning modes:
\begin{equation}
\label{eq:loss_mse}
\mathcal{L}_{\text{MSE}} = \frac{1}{2} \left( \text{MSE}(\hat{y}_{\text{SC}}, y_{\text{SC}}) + \text{MSE}(\hat{y}_{\text{LC}}, y_{\text{LC}}) \right) .
\end{equation}

However, since the switching decision depends on the predicted margin $\hat{y}_{\text{LC}} - \hat{y}_{\text{SC}}$ (Equation (\ref{eq:decision_rule})), this objective alone does not directly supervise the decision signal. To address this, we introduce a margin loss ($\mathcal{L}_{\text{margin}}$) that penalizes the error in the predicted difference:
\begin{equation}
\label{eq:loss_margin}
\mathcal{L}_{\text{margin}} = \text{MSE}\left( \hat{y}_{\text{LC}} - \hat{y}_{\text{SC}},\ y_{\text{LC}} - y_{\text{SC}} \right) .
\end{equation}

The final training objective is the sum of the standard MSE loss and the margin loss ($\mathcal{L}_{\text{switch}}$):
\begin{equation}
\label{eq:loss_switch}
\mathcal{L}_{\text{switch}} = \mathcal{L}_{\text{MSE}} + \lambda_{\text{margin}} \cdot \mathcal{L}_{\text{margin}} ,
\end{equation}
where $\lambda_{\text{margin}} $ is a tunable hyperparameter controlling the weight of margin supervision.

This formulation enhances decision supervision by directly aligning the training objective with the switcher's routing mechanism. In practice, it results in more consistent pathway selection and improved overall performance. The impact of the margin loss \(\mathcal{L}_{\text{margin}}\) is further examined in Section~\ref{subsec:effectiveness_modified_loss}.

\section{Experiments}

\subsection{Experimental Setup}
\label{sec:exp_setup}

\paragraph{Models}
We evaluate our ThinkSwitcher framework on three open-source reasoning models from the DeepSeek-R1 series: DeepSeek-R1-Distill-Qwen-1.5B, DeepSeek-R1-Distill-Qwen-7B, and DeepSeek-R1-Distill-Qwen-14B. These models are distilled versions of DeepSeek-R1 and fine-tuned from Qwen2.5-Math-1.5B, Qwen2.5-Math-7B, and Qwen2.5-14B respectively. All three exhibit strong performance on complex reasoning tasks for their respective scales and serve as representative backbones for our evaluation.
\vspace{-0.0cm}

\paragraph{Training Data} 
To train ThinkSwitcher, we construct a comprehensive and representative dataset by aggregating the training sets of several well-established math benchmarks, spanning a wide range of reasoning difficulties. Specifically, our training data combines several sources: we use the training splits from MATH~\cite{hendrycks2measuring} and GSM8K~\cite{cobbe2021training}. The data also includes the historical AIME problems from 1983 to 2023~\cite{AIME, aime_1983_2024}. Finally, a subset from Omni-MATH~\cite{gao2025omnimath} is incorporated, which consists of approximately 3,900 problems obtained by excluding the Omni-MATH-500 test set used for evaluation.
There is no overlap between the data used to train ThinkSwitcher and any of the test sets used in our experiments. Detailed training hyperparameters and implementation specifics are provided in Appendix \ref{appendix:implementation_details}.

\vspace{-0.0cm}
\paragraph{Evaluation}
We evaluate ThinkSwitcher on a diverse set of datasets spanning three difficulty levels: (1) \textbf{Basic}, represented by GSM8K~\cite{cobbe2021training}; (2) \textbf{Intermediate}, using MATH-500, a subset of MATH~\cite{hendrycks2measuring}; and (3) \textbf{Competition}, including AIME~\cite{AIME}, LiveAoPSBench~\cite{mahdavi2025leveragingonlineolympiadlevelmath}, Omni-MATH-500~\cite{gao2025omnimath}, and the math subset of OlympiadBench~\cite{he2024olympiadbenchchallengingbenchmarkpromoting}. Detailed descriptions are provided in Appendix~\ref{appendix:datasets_details}.

\begin{table*}[!ht]
\centering
\resizebox{\textwidth}{!}{%
\begin{tabular}{lcccccccccccccccc}
\toprule
\multirow{2}{*}{\textbf{Method}} & \multicolumn{2}{c}{\textbf{GSM8K}} & \multicolumn{2}{c}{\textbf{MATH500}} & \multicolumn{2}{c}{\textbf{AIME24}} & \multicolumn{2}{c}{\textbf{AIME25}} & \multicolumn{2}{c}{\textbf{LiveAoPS}} & \multicolumn{2}{c}{\textbf{OmniMATH}} & \multicolumn{2}{c}{\textbf{OlymBench}} & \multicolumn{2}{c}{\textbf{Avg.}} \\
\cmidrule(lr){2-3} \cmidrule(lr){4-5} \cmidrule(lr){6-7} \cmidrule(lr){8-9} \cmidrule(lr){10-11} \cmidrule(lr){12-13} \cmidrule(lr){14-15} \cmidrule(lr){16-17}
& Acc. & Tok. & Acc. & Tok. & Acc. & Tok. & Acc. & Tok. & Acc. & Tok. & Acc. & Tok. & Acc. & Tok. & Acc. & Tok. \\
\midrule
\multicolumn{17}{c}{\textbf{DeepSeek-R1-Distill-Qwen-1.5B}} \\
\midrule
SC-Only & 71.7 & 246 & 68.2 & 880 & 12.5 & 4124 & 14.2 & 3808 & 35.0 & 2028 & 28.3 & 1949 & 32.7 & 1805 & 37.5 & 2120 \\
LC-Only & 85.5 & 2335 & 83.2 & 4907 & 28.3 & 12275 & 29.2 & 11235 & 46.5 & 10195 & 37.4 & 10613 & 43.7 & 9251 & 50.5 & 8687 \\
Random (50/50) & 78.7 & 1281 & 76.5 & 2821 & 18.3 & 8913 & 22.5 & 7187 & 40.3 & 6055 & 33.3 & 6224 & 38.5 & 5663 & 44.0 & 5449 \\
Random (25/75) & 82.2 & 1812 & 80.5 & 3838 & 24.6 & 10517 & 25.0 & 10186 & 42.9 & 8273 & 35.1 & 8265 & 40.9 & 7468 & 47.3 & 7194 \\
Random (10/90) & 84.5 & 2124 & 82.4 & 4465 & 25.4 & 11534 & 25.8 & 10780 & 44.7 & 9560 & 36.9 & 9582 & 42.9 & 8616 & 49.0 & 8094 \\
\midrule
BERT & 80.6 & \textbf{1424} & 80.1 & \textbf{3743} & 18.8 & \textbf{6872} & 25.0 & \textbf{6438} & 40.4 & \textbf{4534} & 31.5 & \textbf{3995} & 38.4 & \textbf{3766} & 45.0 & \textbf{4396} \\

\rowcolor{mylightblue} ThinkSwitcher & \textbf{84.7} & 2114 & \textbf{82.4} & 4544 & \textbf{23.3} & 8192 & \textbf{28.3} & 6689 & \textbf{43.9} & 7010 & \textbf{35.3} & 6238 & \textbf{43.1} & 5831 & \textbf{48.7} & 5803 \\
\midrule
\multicolumn{17}{c}{\textbf{DeepSeek-R1-Distill-Qwen-7B}} \\
\midrule
SC-Only & 87.5 & 257 & 78.9 & 617 & 20.8 & 1781 & 18.3 & 1548 & 39.4 & 1171 & 35.6 & 1135 & 41.2 & 1044 & 46.0 & 1079 \\
LC-Only & 93.0 & 1672 & 91.7 & 3828 & 51.7 & 10884 & 39.2 & 11214 & 64.3 & 8592 & 54.1 & 9399 & 58.5 & 7728 & 64.6 & 7617 \\
Random (50/50) & 90.4 & 962 & 84.9 & 2244 & 39.6 & 6687 & 23.3 & 4620 & 52.8 & 5137 & 44.9 & 4948 & 50.5 & 4244 & 55.2 & 4120 \\
Random (25/75) & 91.7 & 1340 & 87.6 & 2963 & 45.0 & 9445 & 25.8 & 7733 & 57.9 & 6989 & 49.0 & 6985 & 55.2 & 5973 & 58.9 & 5918 \\
Random (10/90) & 92.6 & 1536 & 90.4 & 3509 & 47.5 & 10274 & 30.8 & 9169 & 61.9 & 7964 & 51.7 & 8396 & 57.3 & 7073 & 61.8 & 6846 \\
\midrule
BERT & 87.9 & \textbf{314} & 89.5 & \textbf{2686} & 45.4 & 9985 & \textbf{39.2} & 11214 & 
57.5 & \textbf{5797} & 50.0 & 7129 & 55.4 & \textbf{5024} & 60.7 & 6021 \\

\rowcolor{mylightblue} ThinkSwitcher & \textbf{92.5} & 1389 & \textbf{91.3} & 3495 & \textbf{48.3} & \textbf{7936} & 37.5 & \textbf{6955} & \textbf{61.6} & 6571 & \textbf{51.2} & \textbf{6345} & \textbf{57.0} & 5147 & \textbf{62.8} & \textbf{5405} \\
\midrule
\multicolumn{17}{c}{\textbf{DeepSeek-R1-Distill-Qwen-14B}} \\
\midrule
SC-Only & 90.7 & 248 & 76.9 & 672 & 18.3 & 2420 & 18.3 & 1593 & 43.6 & 1404 & 34.5 & 1386 & 40.4 & 1269 & 46.1 & 1284 \\
LC-Only & 95.3 & 1512 & 92.8 & 3644 & 61.7 & 9887 & 42.5 & 11081 & 68.5 & 7843 & 59.2 & 8619 & 60.0 & 7097 & 68.6 & 7098 \\
Random (50/50) & 93.1 & 894 & 84.3 & 2066 & 42.1 & 5316 & 24.2 & 5866 & 55.7 & 4619 & 47.3 & 5157 & 50.2 & 4068 & 56.7 & 3998 \\
Random (25/75) & 94.2 & 1213 & 88.1 & 2809 & 49.2 & 6764 & 30.0 & 6459 & 62.3 & 6270 & 53.7 & 7018 & 56.0 & 5563 & 61.9 & 5157 \\
Random (10/90) & 95.1 & 1395 & 90.9 & 3262 & 53.8 & 9044 & 40.0 & 9353 & 66.8 & 7304 & 56.6 & 7903 & 58.4 & 6530 & 65.9 & 6399 \\
\midrule
BERT & 91.3 & \textbf{303} & 88.3 & \textbf{2445} & 57.1 & 9568 & \textbf{42.5} & 11081 & 63.5 & 6199 & \textbf{55.1} & 7018 & \textbf{57.6} & 5269 & 65.0 & 5983 \\

\rowcolor{mylightblue} ThinkSwitcher & \textbf{94.3} & 1042 & \textbf{92.7} & 3572 & \textbf{60.4} & \textbf{8044} & \textbf{42.5} & \textbf{10065} & \textbf{65.8} & \textbf{6018} & 54.9 & \textbf{5828} & \textbf{57.6} & \textbf{4651} & \textbf{66.9} & \textbf{5603} \\
\bottomrule
\end{tabular}
}
\caption{
Performance of ThinkSwitcher and baseline methods across three model sizes (1.5B, 7B, and 14B) in the \textbf{DeepSeek-R1-Distill-Qwen} series, evaluated on diverse math benchmarks. `Avg.'' shows macro-averaged accuracy and token counts across the seven benchmarks. \textbf{Bold} numbers indicate performance surpassing the BERT baseline
}
\label{tab:main_results_combined}
\vspace{-0.2cm}
\end{table*}

\vspace{-0.1cm}
\paragraph{Baselines}
\label{sec:baselines}
We compare the performance of ThinkSwitcher against four baselines, each representing a distinct reasoning strategy. \textbf{SC-Only} applies our prompting method to induce short CoT on all problems. This method prioritizes efficiency at the expense of reasoning performance on more complex tasks. \textbf{LC-Only} reflects the default behavior of LRMs, which follows long CoT without intervention and typically achieves higher accuracy on complex tasks. The \textbf{Random} switcher selects between short and long CoT prompts for each input based on predefined probabilities; specifically, ``Random (x/y)'' indicates that the long CoT mode is selected with a probability of x\%, and the short CoT mode with a probability of y\%. This serves as a stochastic control to assess whether the learned switcher offers improvements over naive selection. Finally, we include a \textbf{BERT}-based switcher. For this baseline, we trained a classifier based on an advanced BERT variant, ModernBERT-base~\cite{warner2024smarterbetterfasterlonger}. This classifier, trained on the same data as ThinkSwitcher, learns to select the reasoning mode likely to yield a higher pass rate and exemplifies a conventional routing method.

\paragraph{Efficiency Metrics}
We measure computational cost as the average number of tokens generated per question. For ThinkSwitcher, this depends on the reasoning path (SC or LC) selected at each question, while for static baselines it reflects the token usage of the fixed prompting strategy.

To evaluate the overall trade-off performance, we use the \textbf{AUC-AC} to quantify how well ThinkSwitcher maintains accuracy across its spectrum of token efficiencies, and its normalized variant \textbf{nAUC-AC} to measure the advantage gained by ThinkSwitcher's adaptive mechanism over a linear interpolation between SC-Only and LC-Only performance points. Detailed definitions of these metrics are provided in Appendix~\ref{app:additional_metrics}.

\vspace{-0.1cm}
\subsection{Overall Results}
\label{sec:main_results}

To assess the effectiveness of ThinkSwitcher, we evaluate its performance across the three previously detailed model scales, analyzing the trade-off between reasoning performance and computational efficiency. For the main results presented in Table~\ref{tab:main_results_combined}, decision thresholds ($\tau$) were selected to achieve a favorable balance between accuracy and efficiency; specifically, $\tau$ is $0.04$ for the 1.5B model, $0.05$ for the 7B model, and $0.03$ for the 14B model. We identify several key insights from these results.

First, ThinkSwitcher achieves a consistent balance between reasoning accuracy and computational cost. Compared to the default LC-Only strategy, which applies full long-form reasoning for every input, ThinkSwitcher reduces the average number of generated tokens by over 20\% while incurring only a slight drop of 1–2\% in average accuracy. This supports our core hypothesis that a single model can reason adaptively without relying on costly, always-on deep CoT processing.

In addition, we also compare ThinkSwitcher against a BERT-based switcher, a conventional classification approach widely used for decision routing. While BERT achieves lower token usage on the 1.5B model, it consistently underperforms ThinkSwitcher in both accuracy and efficiency on the larger models. Even on the 1.5B setting, this token saving comes at the cost of a substantial drop in accuracy, limiting its practical utility. In contrast, ThinkSwitcher delivers stronger performance with fewer parameters and consistently achieves lower token consumption across all model scales, making it a more robust and deployable solution.
\begin{figure*}[!h]
    \centering
    \vspace{-0.2cm}
    \includegraphics[width=0.98\textwidth]{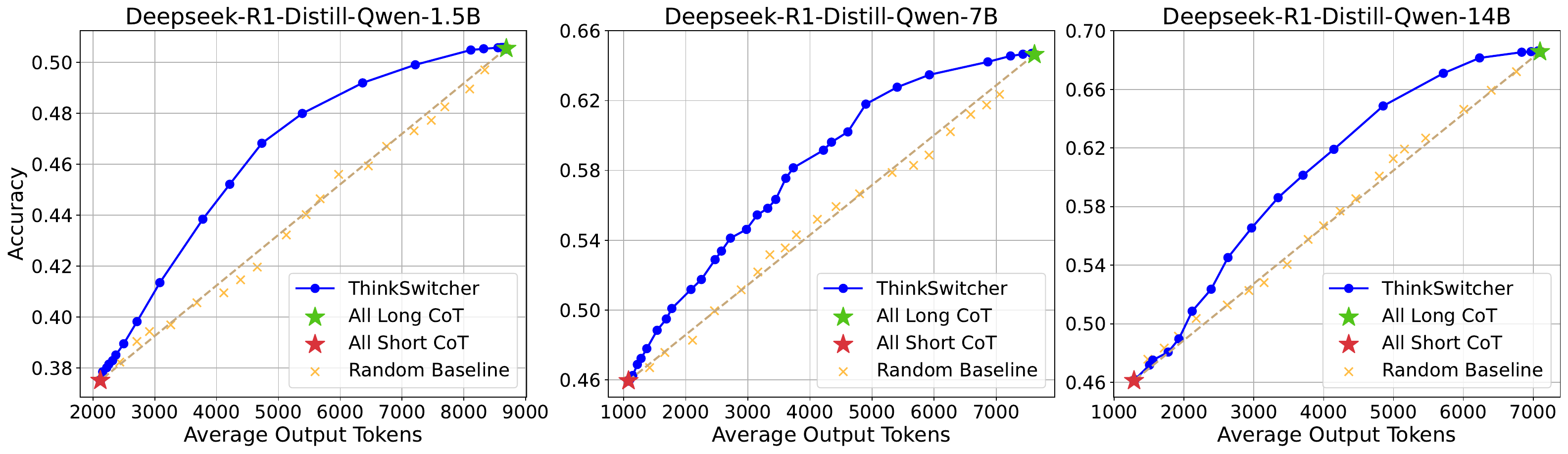}
    \vspace{-0.2cm}
    \caption{
    Trade-off between average accuracy and cost (measured by average output tokens) for the three DeepSeek-R1-Distill-Qwen model sizes. Each point on the ThinkSwitcher curves corresponds to a different $\tau$ value.}
    \label{fig:tradeoff_combined}
\vspace{-0.3cm}
\end{figure*}

Finally, we observe that smaller models benefit disproportionately from adaptive switching. For example, on the 1.5B model, ThinkSwitcher reduces token usage by more than 30\% while maintaining robust accuracy. This suggests that weaker models are more prone to over-elaboration and thus stand to gain more from selective short-form reasoning.

\vspace{-0.0cm}
\section{Analyses}
\vspace{-0.1cm}
\label{sec:analysis}

To gain a deeper understanding of the behavior and effectiveness of ThinkSwitcher, we conduct in-depth analyses. These include examining decision dynamics (Sec~\ref{subsec:scaling_tradeoff}), quantifying the impact of different training objectives (Sec~\ref{subsec:effectiveness_modified_loss}), evaluating key design components (Sec~\ref{subsec:passrate_estimation_analysis}), and assessing the computational efficiency of the switching mechanism (Sec~\ref{subsec:computational_efficiency}). Unless otherwise specified, experiments in this section use the DeepSeek-R1-Distill-Qwen-7B model, selected for its favorable trade-off between efficiency and generality.

\vspace{-0.1cm}
\subsection{Scaling the Trade-off}
\label{subsec:scaling_tradeoff}
In this section, we evaluate the balance between accuracy and cost of ThinkSwitcher across three model scales: DeepSeek-R1-Distill-Qwen-1.5B, 7B, and 14B. For each model, we plot average accuracy against average token cost by systematically sweeping the decision threshold \(\tau\). The resulting trade-off curves are presented in Figure~\ref{fig:tradeoff_combined}.

Across all model sizes, ThinkSwitcher consistently outperforms the random baseline, demonstrating its ability to dynamically select appropriate reasoning modes to balance performance and computational cost. Compared to fixed prompting strategies (SC-only and LC-only), it achieves a stronger Pareto frontier by delivering better accuracy–cost trade-offs across operating points.

The accuracy–cost curves reveal a notable trend: smaller models exhibit trade-off curves positioned closer to the top-left corner, indicating a more favorable balance between performance and efficiency. This observation aligns with discussions in Section~\ref{sec:main_results}, where it was noted that these smaller models (e.g., 1.5B) tend to generate significantly longer responses under LC-only prompting, resulting in higher average token costs. Such behavior suggests that weaker models rely more heavily on extended reasoning to arrive at correct answers and are thus more sensitive to the choice of prompting strategy. Consequently, ThinkSwitcher achieves greater cost savings in these settings by selectively avoiding unnecessarily long reasoning paths.

These findings demonstrate that ThinkSwitcher provides a practical and scalable solution for cost-aware reasoning across models of varying capacity.

\vspace{-0.1cm}
\subsection{Effectiveness of the Switcher Loss}
\label{subsec:effectiveness_modified_loss}

As detailed in Section~\ref{subsec:training_loss}, our final training objective incorporates a differential loss term, \(\mathcal{L}_{\text{margin}}\), which explicitly supervises the predicted performance gap between CoT reasoning pathways. This additional supervision is critical for improving the quality of switcher decisions, as the routing mechanism is defined by Equation~(\ref{eq:decision_rule}) and depends directly on the predicted margin \(\hat{y}_{\text{LC}} - \hat{y}_{\text{SC}}\).

To examine its impact, we conducted an ablation study by varying the weight \(\lambda_{\text{margin}}\) in the combined loss defined in Equation~(\ref{eq:loss_switch}). For the metric, we use normalized AUC-AC (nAUC-AC), which reflects the overall cost–accuracy trade-off relative to a linear baseline. The results are shown in Table~\ref{tab:impact_lambda_margin}.

\begin{table}[h]
\centering
\small
\begin{tabular}{lc}
\toprule
\textbf{Margin Loss Weight (\(\lambda_{\text{margin}}\))} & \textbf{nAUC-AC} \\
\midrule
0 (No Margin Loss) & 167 \\
1 (Moderate Weight) & \textbf{199} \\
2 (High Weight) & 166 \\
\bottomrule
\end{tabular}
\caption{Effect of margin loss weight (\(\lambda_{\text{margin}}\)) on downstream task performance, measured by nAUC-AC.}
\label{tab:impact_lambda_margin}
\vspace{-0.2cm}
\end{table}

The results show that incorporating margin supervision with a moderate weight (\(\lambda_{\text{margin}} = 1\)) yields the best overall performance. This confirms the utility of explicitly aligning the training objective with the decision criterion. However, overly emphasizing the margin term (\(\lambda_{\text{margin}} = 2\)) degrades performance, suggesting that maintaining a balance between absolute accuracy and margin accuracy is essential for reliable switching.

\vspace{-0.1cm}
\subsection{Impact of Pass Rate Estimation Quality}
\label{subsec:passrate_estimation_analysis}

\begin{figure}[!t]
\centering
\vspace{-0.35cm}
\includegraphics[width=0.4\textwidth]{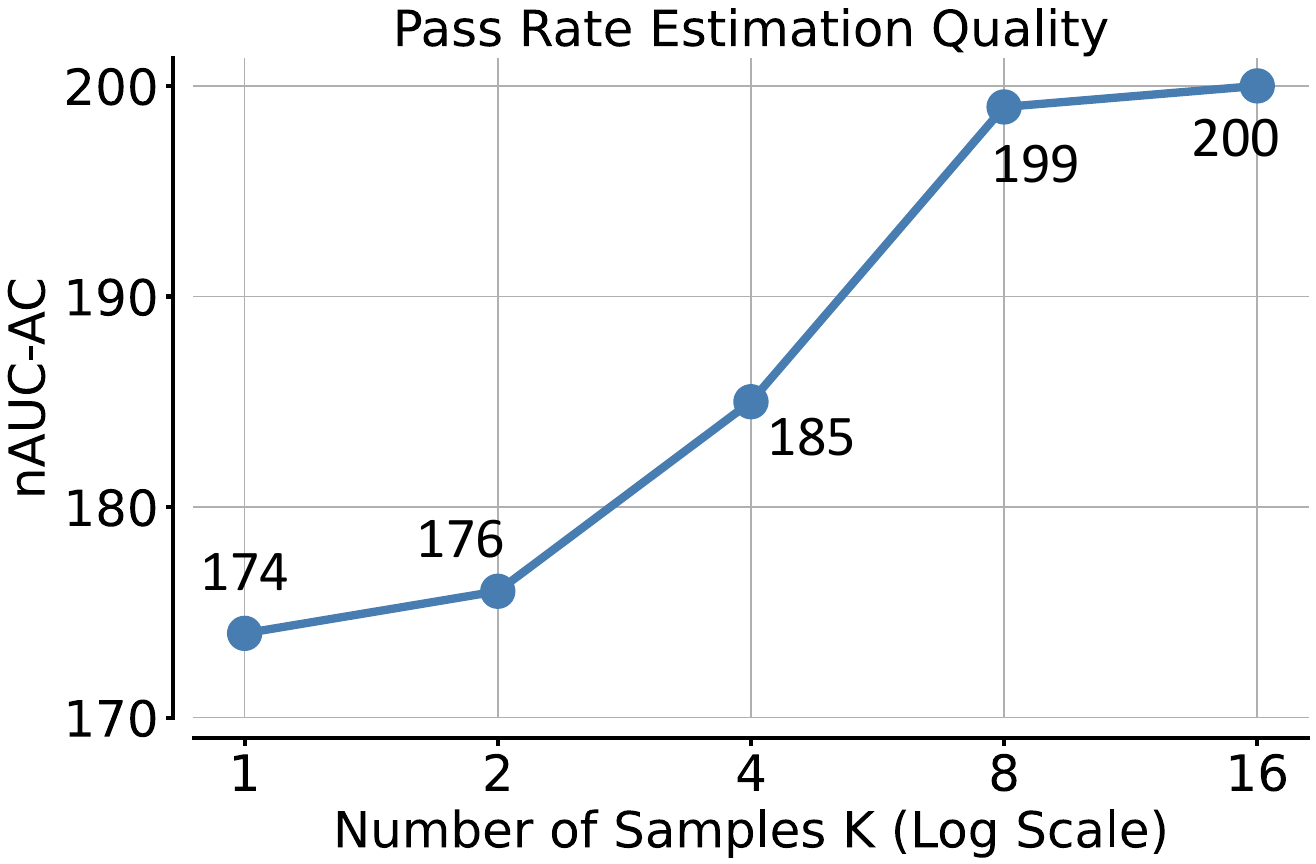}
\vspace{-0.2cm}
\caption{Performance with different values of $k$ used to estimate pass rates in training data construction.}
\label{fig:impact_k_sampling}
\vspace{-0.45cm}
\end{figure}

The switcher is trained using pass rates estimated over $k$ sampled responses per pathway (SC and LC) for each training instance. To assess how the estimation quality of these supervisory signals affects downstream performance, we vary $k \in \{1, 2, 4, 8, 16\}$ during the training data construction phase and evaluate the resulting switchers within the full ThinkSwitcher framework. Results are shown in Figure~\ref{fig:impact_k_sampling}, measured by nAUC-AC.

We observe a clear upward trend in performance as $k$ increases: nAUC-AC improves from approximately 174 at $k=1$ to around 199 at $k=8$, indicating that higher-quality pass rate estimates provide more reliable training signals and lead to more effective switching policies. However, this trend saturates beyond $k=8$, as increasing to $k=16$ yields only marginal improvement. This suggests diminishing returns, where further reducing the variance of the pass rate estimates no longer translates to meaningful performance gains. We therefore adopt $k=8$ in our main experiments, as it provides a favorable trade-off between supervision fidelity and the computational overhead of constructing the training dataset. This setting ensures stable switcher performance without incurring the high cost of excessive sampling.

\vspace{-0.1cm}
\subsection{Computational Efficiency of Switching}
\label{subsec:computational_efficiency}

To assess the practical efficiency of the proposed switching mechanism, we analyze its impact on inference-time computational cost across different model scales. This evaluation is crucial to ensure that the efficiency gains brought by adaptive reasoning are not offset by the additional overhead introduced by the switcher module.

As shown in Table~\ref{tab:flops_comparison_lc}, the switcher module introduces negligible overhead: its per-query compute cost consistently remains in the range of millions of FLOPs, which is insignificant compared to the trillions required for LLM decoding.
Although the number of switcher parameters varies slightly with model scale, this variation is solely attributed to differences in the hidden state dimensionality of the underlying LRMs, which only affects the input layer of the switcher. The rest of the switcher architecture remains fixed, ensuring the module stays lightweight and suitable for practical deployment.

    \begin{table}[t]
\centering
\small
\vspace{-0.25cm}
\begin{tabular}{@{}lccc@{}}
\toprule
\textbf{Metric} & \textbf{1.5B} & \textbf{7B} & \textbf{14B} \\
\midrule
Switcher Params & 2.89M & 4.98M & 6.56M \\
Switcher FLOPs & 5.77M & 9.97M & 13.11M \\
\midrule
LLM FLOPs (LC-Only) & 26.06T & 106.64T & 198.74T \\
LLM FLOPs (TS) & 17.41T & 75.67T & 156.88T \\
\bottomrule
\end{tabular}
\vspace{-0.15cm}
\caption{
Per-query computational costs across model scales. “Switcher Params” and “Switcher FLOPs” denote the size and compute overhead of the switching module. “LLM FLOPs (LC-Only)” refers to decoding cost under long CoT reasoning, while “LLM FLOPs (TS)” reflects the decoding cost with ThinkSwitcher. 
}
\label{tab:flops_comparison_lc}
\vspace{-0.45cm}
\end{table}

More importantly, switching enables substantial reductions in overall decoding cost. Compared to the LC-Only baseline, which always applies long-form reasoning, adaptive switching reduces LLM decoding FLOPs by 20–30\% across all model sizes. These savings arise from dynamically selecting shorter reasoning paths for simpler queries, thereby improving efficiency without compromising performance. Collectively, these findings validate our switching-based approach as a practical and scalable solution for cost-aware reasoning.

\vspace{-0.2cm}
\section{Conclusion}
\vspace{-0.1cm}

While large reasoning models (LRMs) excel at complex tasks, they often default to unnecessarily long chain-of-thought (CoT) reasoning even for simple problems and lead to significant computational overhead. We introduced ThinkSwitcher, a lightweight and adaptive framework that addresses this \emph{overthinking} tendency by dynamically selecting between short and long CoT modes based on task complexity. ThinkSwitcher harnesses the model’s latent ability for concise reasoning—elicited via prompt design—and integrates a self-supervised switcher module to automate mode selection without modifying the backbone architecture or requiring additional training. Experiments across diverse benchmarks show that ThinkSwitcher reduces computational cost by 20–30\% while preserving strong performance on complex tasks. These findings highlight adaptive reasoning control as a scalable and effective approach to mitigating overthinking in LRMs and supporting efficient unified deployment.

\section*{Limitations}

While our method demonstrates strong empirical performance, it still has several limitations.
First, while the ThinkSwitcher framework demonstrates strong results on mathematical reasoning benchmarks, its applicability to other complex reasoning tasks such as code generation has not yet been explored.
Second, due to computational constraints, our experiments are limited to models with up to 14B parameters. Nonetheless, the method is inherently scalable, and we expect it to extend effectively to larger or architecturally diverse models, which we leave for future investigation.

\bibliography{main}

\appendix

\section{Induced Short CoT Performs on Challenging Tasks}
\label{appendix:additional_shortcot_results}

This section extends the analysis in Section~\ref{subsec:generality_efficiency_performance} by evaluating the performance of induced short CoT reasoning on additional two high-difficulty benchmarks: AIME24 and AIME25.

\begin{figure*}[!h]
\centering
\includegraphics[width=0.95\textwidth]{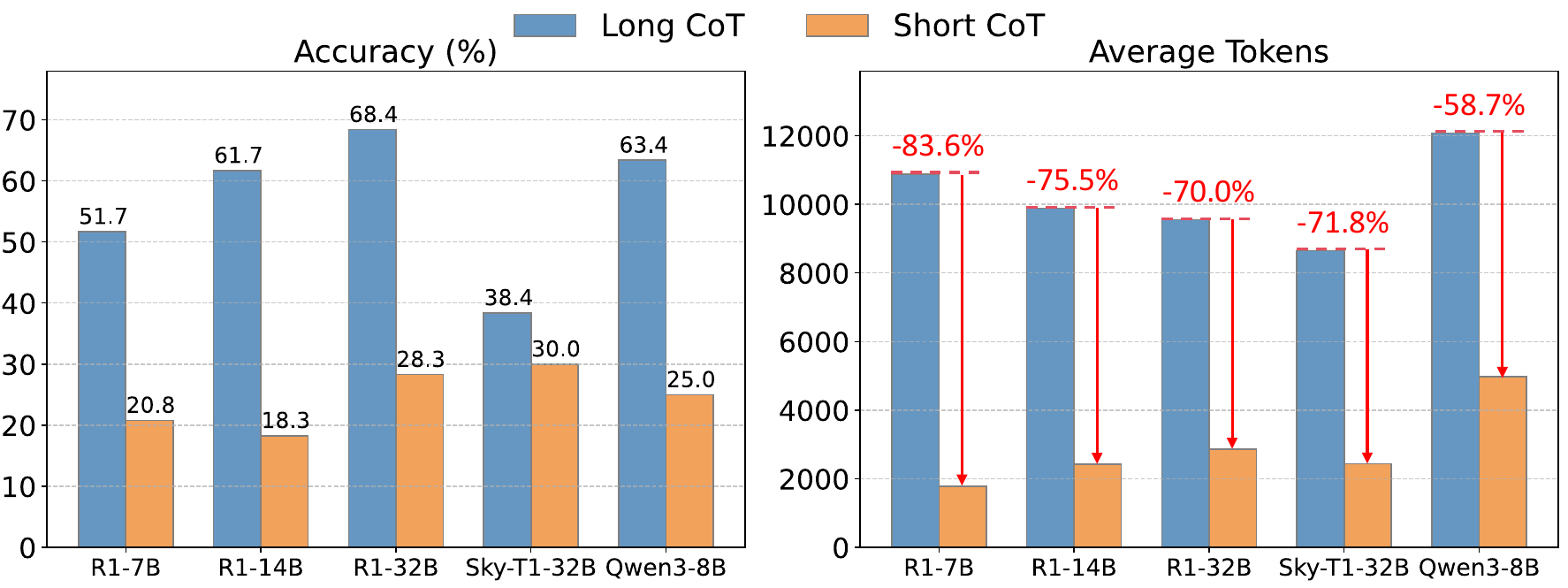}
\caption{Comparison of long CoT and induced short CoT on the AIME24.}
\label{fig:aime24_short_cot_performance}
\end{figure*}

\begin{figure*}[!h]
\centering
\includegraphics[width=0.95\textwidth]{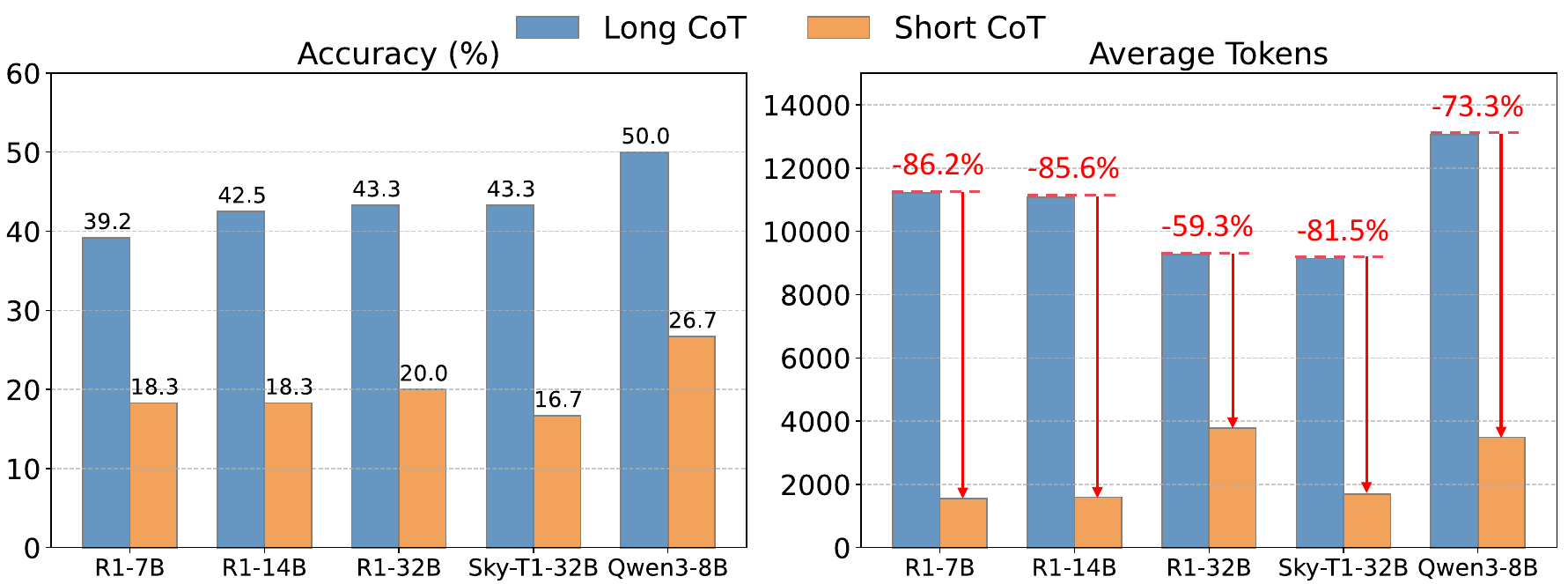}
\caption{Comparison of long CoT and induced short CoT on the AIME25.}
\label{fig:aime25_short_cot_performance}
\end{figure*}

As shown in Figures~\ref{fig:aime24_short_cot_performance} and \ref{fig:aime25_short_cot_performance}, induced short CoT consistently yields substantial reductions in token usage on these high-difficulty AIME benchmarks. Aligning with our observations in the main paper, while short CoT exhibits an accuracy drop on such particularly complex problems, it still retains problem-solving capabilities. These AIME results therefore confirm that the significant efficiency benefits of short CoT induction extend to competition-level tasks. This reinforces its utility as an efficient reasoning mode, particularly when long CoT can be strategically invoked for instances requiring maximum performance.

\section{Templates for Short and Long CoT}
\label{appendix:prompt_templates}

This section presents the prompt templates used to induce long and short CoT reasoning modes from LRMs. Each template contains a placeholder \texttt{\{question\}} for inserting the input query and includes instructions for the model to reason step by step, with the final answer enclosed in \texttt{\textbackslash boxed\{\}}.

\begin{figure}[H]
\centering
\includegraphics[width=0.48\textwidth]{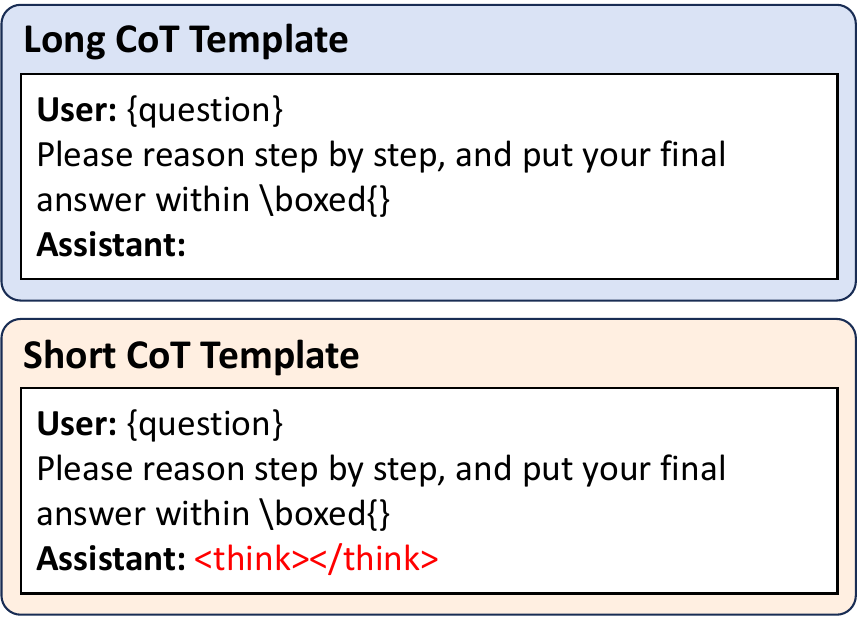}
\caption{Long and short CoT prompt template.}
\label{fig:prompt_template}
\vspace{-0.4cm}
\end{figure}

As shown in Figure~\ref{fig:prompt_template}, two templates differ only in the inclusion of an empty \texttt{<think>} block in the short CoT version, which reliably triggers concise reasoning behavior. As discussed in Section~\ref{subsec:core_observation}, this minimal intervention is sufficient to suppress unnecessary elaboration and induce efficient CoT responses without modifying the model itself.

\section{Mechanism Behind Short CoT Induction}
\label{appendix:mechanism_behind_short_cot}

What enables simple prompts to reliably induce short chains of thought in LRMs? We hypothesize that this behavior arises from the suppression or bypassing of the elaborate reasoning processes and stylistic conventions instilled during reasoning-specific fine-tuning.
When such patterns are deactivated or inhibited, the model appears to revert to a generation mode that more closely resembles its pre-finetuning behavior, favoring concise and direct outputs over elaborated reasoning.

To investigate this hypothesis, we conducted a semantic similarity analysis across three categories of outputs on the MATH500 benchmark \cite{lightman2024lets}. 
We used the \texttt{all-mpnet-base-v2} model\footnote{\url{https://huggingface.co/sentence-transformers/all-mpnet-base-v2}} to compute sentence embeddings for the reasoning outputs generated under different prompting conditions, and calculated cosine similarity between the following categories:
\begin{enumerate}[leftmargin=0.6cm]
    \item \textbf{ISC (Induced Short CoT):} Responses generated by DeepSeek-R1-Distill-Qwen-7B using our short CoT induction strategy.
    \item \textbf{LCS (Long CoT Summary):} The final summary portion generated after the \texttt{</think>} token in DeepSeek-R1-Distill-Qwen-7B's default long CoT setting, typically containing the conclusion or final answer.
    \item \textbf{OC (Original CoT):} Natural CoT responses generated by Qwen2.5-Math-7B (the base model for DeepSeek-R1-Distill-Qwen-7B), which is optimized for short CoT reasoning.
\end{enumerate}

\begin{table}[h]
\centering
\vspace{-0.2cm}
\begin{tabular}{lc}
\toprule
Pairwise Comparison & Cosine Similarity \\
\midrule
ISC vs. OC          & 0.926 \\
ISC vs. LCS         & 0.919 \\
LCS vs. OC          & 0.916 \\
\bottomrule
\end{tabular}
\caption{Cosine similarity between response embeddings from ISC, LCS, and OC categories.}
\label{tab:similarity_scores}
\vspace{-0.2cm}
\end{table}

The cosine similarity scores are reported in Table~\ref{tab:similarity_scores}. Based on these values, we observe the following ranking:
\[
\text{Sm(ISC, OC)} > \text{Sm(ISC, LCS)} > \text{Sm(LCS, OC)}
\]

This ranking supports our suppression-and-reversion hypothesis. ISC responses exhibit greater semantic similarity to OC outputs from the general-purpose reasoning model than to LCS segments produced by the LRM under its long-form reasoning mode. This suggests that short CoT prompting may suppress fine-tuned long-form reasoning patterns, allowing the model to revert to a more concise reasoning style latent in earlier training. These results highlight the flexible behavioral priors retained by LRMs, which can be selectively activated or suppressed through external prompts.

\section{Implementation Details}
\label{appendix:implementation_details}

The switcher module is a Multi-Layer Perceptron (MLP) consisting of 5 linear layers with hidden dimensions [1024, 768, 512, 256, 2]. The architecture utilizes Rectified Linear Unit (ReLU) activation functions \cite{agarap2019deeplearningusingrectified}, Batch Normalization \cite{ioffe2015batchnorm}, and Dropout \cite{srivastava2014dropout}.

For training the switcher, the AdamW optimizer \cite{loshchilov2019decoupledweightdecayregularization} was employed. Hyperparameter optimization was performed using Ray Tune \cite{liaw2018tuneresearchplatformdistributed}. The learning rate was selected through a search over a log-uniform distribution in the range [$1 \times 10^{-5}$, $1 \times 10^{-2}$]. The batch size was chosen from the set \{16, 32, 64, 128\}. The dropout rate was explored over a uniform distribution in the interval [0.0, 0.5]. Training was conducted for a maximum of 50 epochs, employing an early stopping strategy based on validation set performance; the model from the best-performing epoch was retained.

The vLLM library \cite{vllm} was utilized to accelerate the sampling of responses from the backbone LRM for data construction. Evaluation of downstream task performance, which informed both the generation of switcher training labels and the final assessment of the ThinkSwitcher framework, was conducted using a customized evaluation framework from the Qwen2.5-math GitHub repository \footnote{\url{https://github.com/QwenLM/Qwen2.5-Math}}.

All experiments were conducted on a computing setup equipped with 4x NVIDIA RTX 4090 GPUs.

\section{Datasets Details}
\label{appendix:datasets_details}

This section describes the evaluation datasets used to assess the performance and generalization of ThinkSwitcher across varying levels of difficulty.

\paragraph{GSM8K~\cite{cobbe2021training}} A dataset of approximately 8,500 high-quality and linguistically diverse grade school math word problems. These problems typically require 2 to 8 steps to solve, and their solutions are written by human problem-solvers, primarily testing multi-step reasoning abilities. The problems cover various arithmetic operations and fundamental concepts taught at the elementary school level.

\paragraph{MATH~\cite{hendrycks2measuring}} Comprises 12,500 challenging mathematics problems, with topics including Algebra, Number Theory, Counting \& Probability, and Geometry, representing a wide range of difficulties up to the high school level. For our evaluation, we utilize a subset of 500 problems, referred to as \textbf{MATH-500}.

\paragraph{AIME~\cite{AIME}} A significant component of the American Mathematics Competitions (AMC) program, positioned in difficulty between the AMC 10/12 and the USAMO (United States of America Mathematical Olympiad). It typically consists of 15 problems to be solved in 3 hours, with answers being integers from 0 to 999. These problems demand deeper mathematical knowledge and creative problem-solving strategies. The AIME dataset used in this evaluation includes problems from the most recent AIME2024 and AIME2025 to cover the latest competition content and challenge levels.

\vspace{-0.1cm}
\paragraph{LiveAoPSBench~\cite{mahdavi2025leveragingonlineolympiadlevelmath}} A dataset of real-time, competition-level mathematics problems collected from online Olympiad mathematics communities, such as the Art of Problem Solving (AoPS) forums. These problems are often proposed by community members or originate from recent minor competitions, possessing strong timeliness and challenge. The subset of LiveAoPSBench used in this evaluation specifically refers to problems that appeared between August 2024 and December 2024, reflecting current trends and the high difficulty of Olympiad mathematics.

\vspace{-0.1cm}
\paragraph{Omni-MATH~\cite{gao2025omnimath}}
A comprehensive benchmark of Olympiad-level mathematics problems designed for universal LLM evaluation. It meticulously curates problems from diverse sources such as national/international Olympiads, training materials, and online forums, covering algebra, geometry, number theory, and combinatorics. Omni-MATH emphasizes problem diversity, fine-grained difficulty scaling, and depth, testing sophisticated reasoning and insight on complex, non-standard tasks. For our evaluation, we utilize a subset of 500 problems from this dataset, referred to as \textbf{Omni-MATH-500}.

\vspace{-0.1cm}
\paragraph{OlympiadBench (math subset)~\cite{he2024olympiadbenchchallengingbenchmarkpromoting}} A challenging, multilingual benchmark designed to evaluate advanced scientific problem-solving in language models, encompassing disciplines like mathematics and physics. We utilize its mathematics subset, which comprises Olympiad-level problems sourced from prestigious international and national competitions, often structured as progressive tasks with interlinked sub-questions. These problems, provided with step-by-step solutions, demand deep conceptual understanding and innovative multi-step reasoning, rigorously testing mathematical capabilities at a high difficulty ceiling.

\section{AUC-AC and nAUC-AC: Metrics for Reasoning Efficiency}
\label{app:additional_metrics}

To systematically evaluate the trade-off between reasoning accuracy and computational cost, we adopt two metrics: the \textbf{Area Under the Accuracy–Cost Curve (AUC-AC)} and its normalized counterpart, \textbf{nAUC-AC}. These metrics quantify how effectively a model balances performance and efficiency across varying output lengths.

Let $A_{TS}(t)$ denote the accuracy achieved at an average token cost $t$. This function is empirically constructed by sweeping the decision threshold $\tau$ to obtain a set of operating points $\{(T(\tau_i), A(\tau_i))\}$. Let $T_{SC}$ and $T_{LC}$ denote the average output tokens of the SC-Only and LC-Only baselines, respectively. The AUC-AC is then defined as the integral of $A_{TS}(t)$ over the interval $[T_{SC}, T_{LC}]$:
\begin{equation}
\label{eq:app_auc_ac}
\text{AUC-AC} = \int_{T_{SC}}^{T_{LC}} A_{TS}(t) \, dt .
\end{equation}

A higher AUC-AC indicates better overall efficiency, reflecting the model’s ability to maintain strong performance under stricter token budgets.

To assess the relative efficiency gain over static baselines, we define the \textbf{Normalized AUC-AC (nAUC-AC)}, which measures the gain over a linear interpolation baseline. 
This baseline corresponds to the linear interpolation between $(T_{SC}, A_{SC})$ and $(T_{LC}, A_{LC})$, representing the expected performance of strategies that use a fixed probability to mix SC-Only and LC-Only modes.
The area under this baseline, denoted $\text{AUC-AC}_{\text{LB}}$, corresponds to the area of the trapezoid:
\begin{equation}
\label{eq:app_auc_ac_lb}
\text{AUC-AC}_{\text{LB}} = \frac{A_{SC} + A_{LC}}{2} \cdot (T_{LC} - T_{SC}) .
\end{equation}

The nAUC-AC is computed as the absolute gain over this linear reference:
\begin{equation}
\label{eq:app_eauc_ac}
\text{nAUC-AC} = \text{AUC-AC}_{\text{TS}} - \text{AUC-AC}_{\text{LB}} .
\end{equation}

This difference captures the extent to which the model’s reasoning efficiency exceeds what can be achieved by linear mixtures of static strategies.

\section{Per-Benchmark Trade-off Results}
\label{app:tradeoff_by_dataset}

\begin{figure*}[htb]
    \centering
    % Subfigure for GSM8K
    \begin{subfigure}{0.98\textwidth} % Adjust width as needed, 0.8 is an example if you want some margin
        \centering
        % Replace with your actual plot file
        \includegraphics[width=\linewidth]{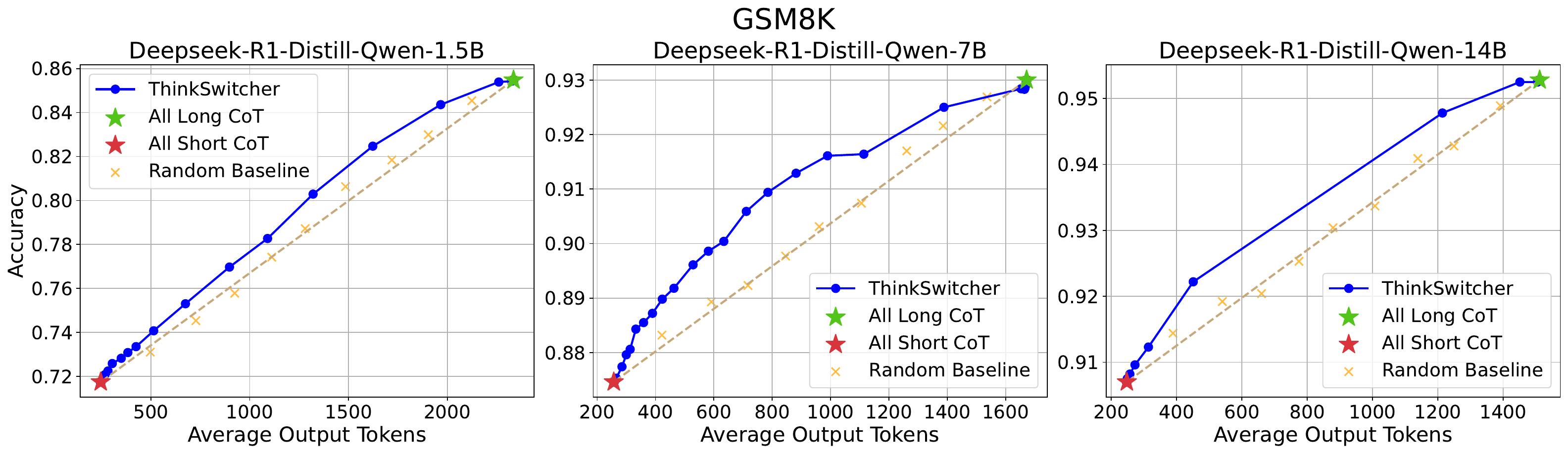} 
        \caption{Trade-off curve for GSM8K.}
        \label{fig:tradeoff_gsm8k}
    \end{subfigure}
    \vfill % Adds some vertical space between subfigures
    % Subfigure for MATH500
       \vspace{0.1cm}
    \begin{subfigure}{0.98\textwidth}
        \centering
        \includegraphics[width=\linewidth]{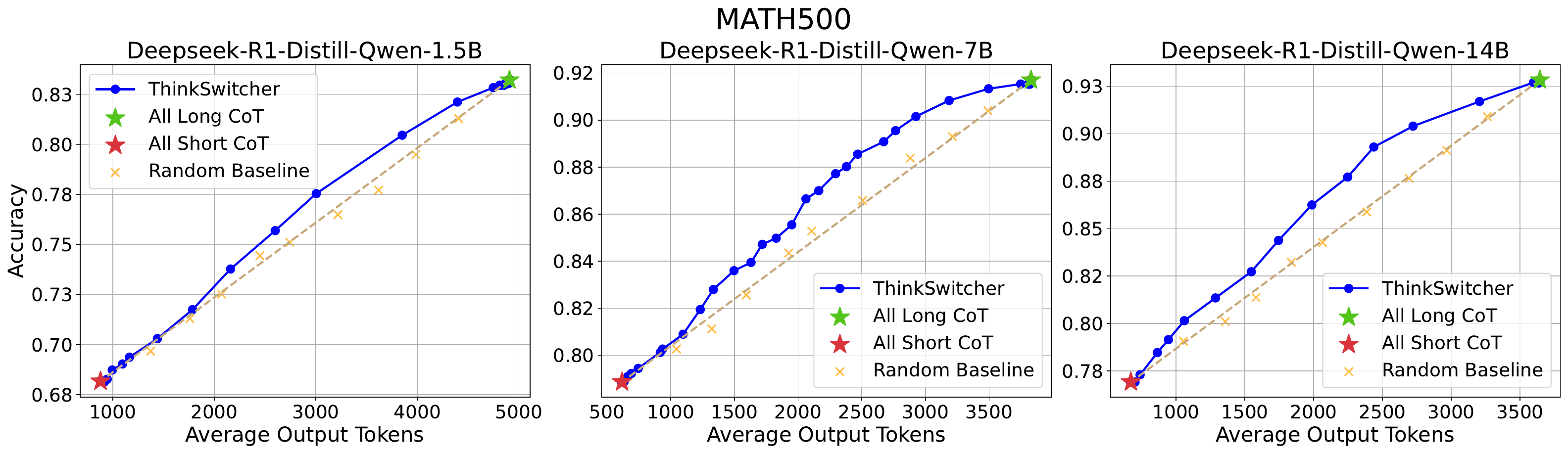}
        \caption{Trade-off curve for MATH500.}
        \label{fig:tradeoff_math500}
    \end{subfigure}
    \vfill
       \vspace{0.1cm}
    % Subfigure for AIME24
    \begin{subfigure}{0.98\textwidth}
        \centering
        \includegraphics[width=\linewidth]{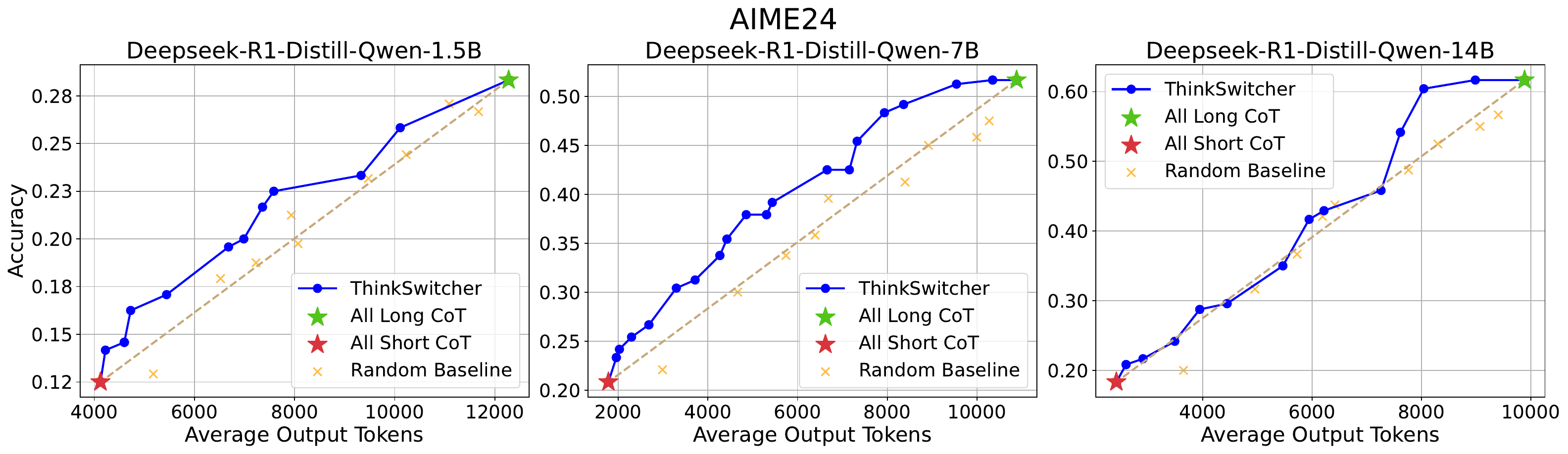}
        \caption{Trade-off curve for AIME24.}
        \label{fig:tradeoff_aime24}
    \end{subfigure}
    \vfill
       \vspace{0.1cm}
    % Subfigure for AIME25
    \begin{subfigure}{0.98\textwidth}
        \centering
        \includegraphics[width=\linewidth]{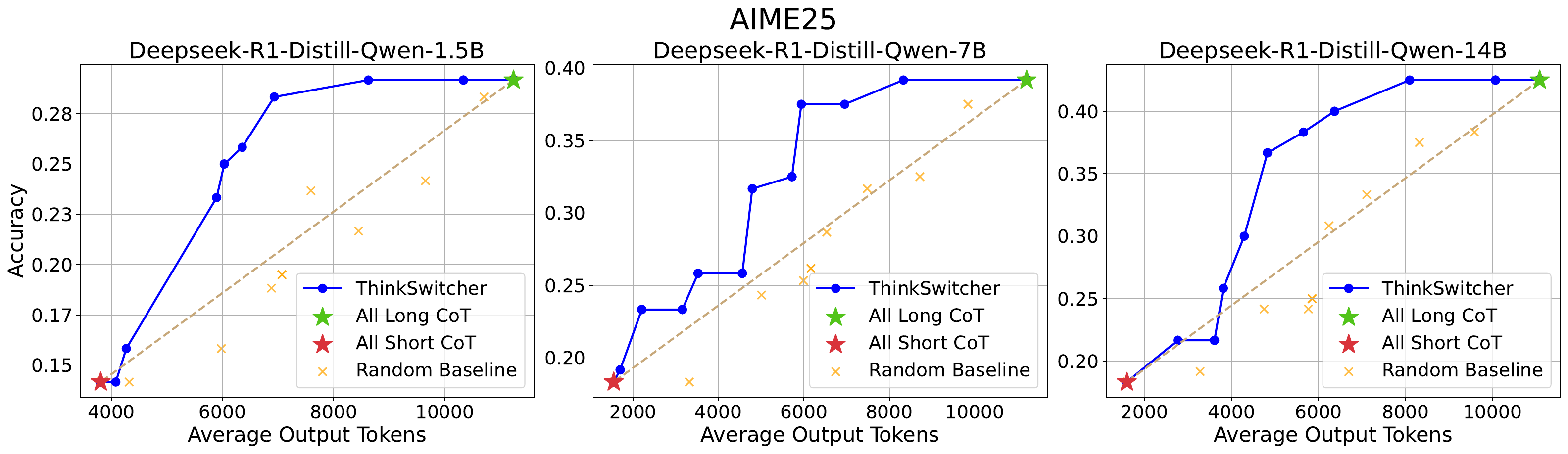}
        \caption{Trade-off curve for AIME25.}
        \label{fig:tradeoff_aime25}
    \end{subfigure}
    \caption{Trade-off between average accuracy and cost for ThinkSwitcher across various datasets (Part 1 of 2).}
    \label{fig:tradeoff_all_datasets_part1}
\end{figure*}

\begin{figure*}[t]
    \centering
    % Subfigure for OmniMATH
    \begin{subfigure}{0.98\textwidth}
        \centering
        \includegraphics[width=\linewidth]{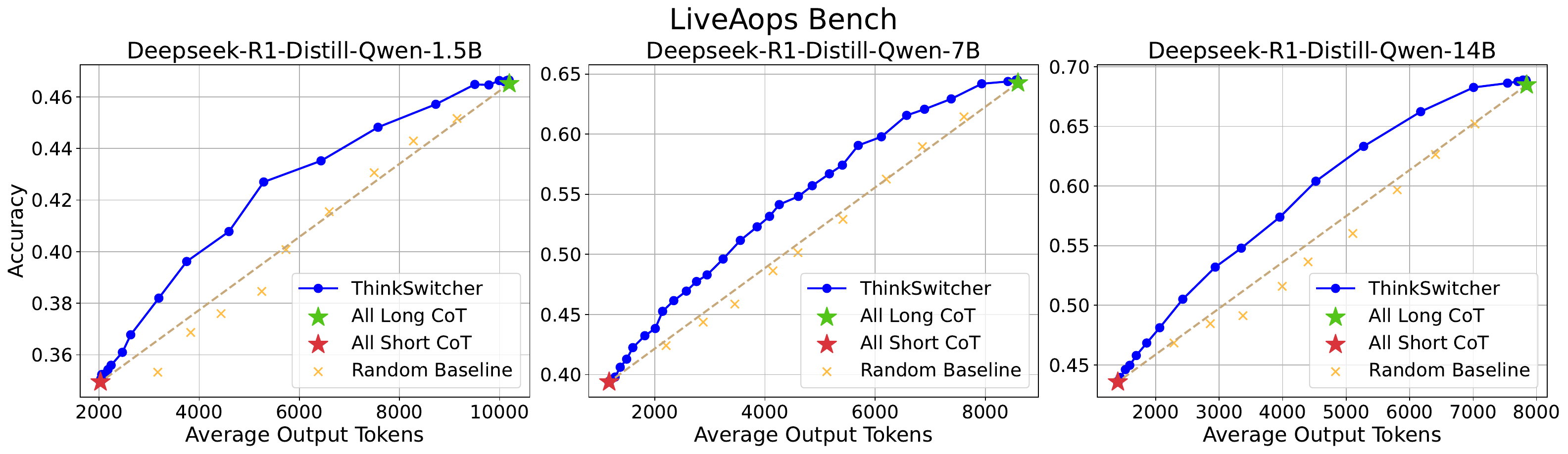}
        \caption{Trade-off curve for LiveAoPS-Bench.}
        \label{fig:tradeoff_liveaops}
    \end{subfigure}
    \vfill
    \vspace{0.1cm}
    \begin{subfigure}{0.98\textwidth}
        \centering
        \includegraphics[width=\linewidth]{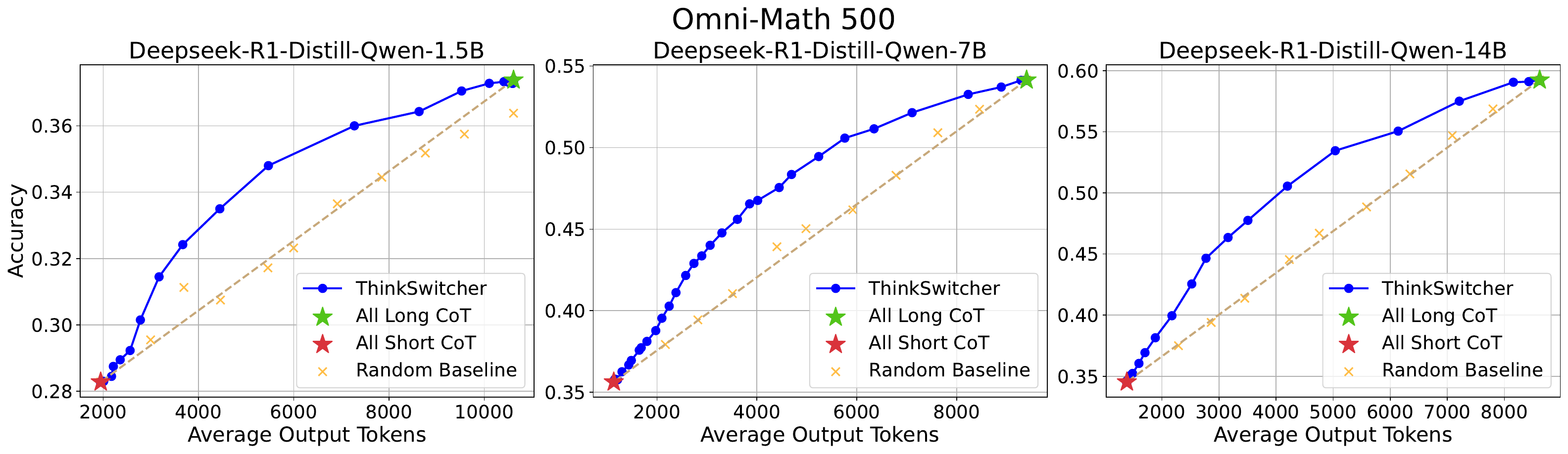}
        \caption{Trade-off curve for Omni-MATH-500.}
        \label{fig:tradeoff_omnimath}
    \end{subfigure}
    \vfill
       \vspace{0.1cm}
    % Subfigure for OlymBench
    \begin{subfigure}{0.98\textwidth}
        \centering
        \includegraphics[width=\linewidth]{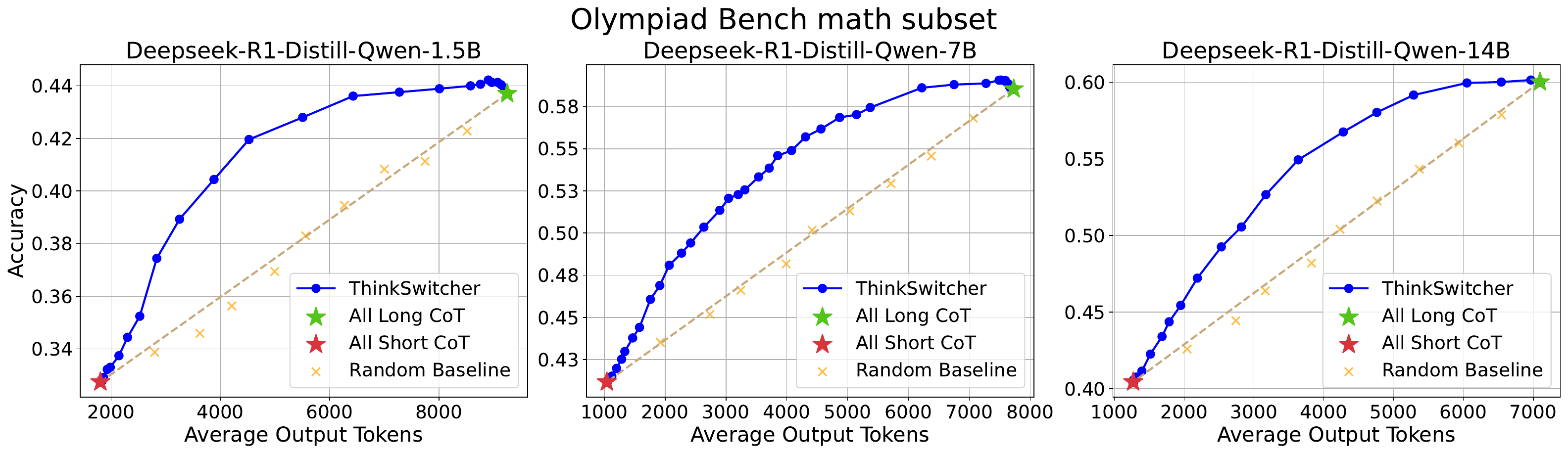}
        \caption{Trade-off curve for OlympiadBench (math subset).}
        \label{fig:tradeoff_olymbench}
    \end{subfigure}
    
    \caption{ Trade-off between average accuracy and cost for ThinkSwitcher across various datasets (Part 2 of 2). }
    \label{fig:tradeoff_all_datasets_part2}
\end{figure*}

This section complements the aggregated analysis in Section~\ref{subsec:scaling_tradeoff} by providing a dataset-level breakdown of the accuracy–efficiency trade-offs. While Figure~\ref{fig:tradeoff_combined} summarizes average performance across all benchmarks and model scales, Figures~\ref{fig:tradeoff_all_datasets_part1} and~\ref{fig:tradeoff_all_datasets_part2} present disaggregated trade-off curves for each of the seven benchmark datasets: GSM8K, MATH500, AIME24, AIME25, LiveAoPS, OmniMATH, and OlymBench.

These disaggregated curves highlight how the effectiveness of ThinkSwitcher varies across task difficulties, offering further insight into when and where dynamic reasoning provides efficiency gains.

\appendix

\end{document}